\documentclass[letterpaper, 10 pt, conference]{ieeeconf} 

\IEEEoverridecommandlockouts                              

\usepackage{amsmath} 
\usepackage{amssymb}  
\usepackage{gensymb}
\usepackage{bm}
\usepackage{cite}

\usepackage{graphicx}
\usepackage{color}
\usepackage{import}
\usepackage{subcaption}
\usepackage{hyperref}
\usepackage{epstopdf}
\usepackage{algorithm}
\usepackage{algpseudocode}
\usepackage{url}
\usepackage{multirow}
\usepackage{multicol}

\newcommand{\argmax}{\mathop{\mathrm{argmax}}\limits}

\title{\LARGE \bf
Consensus-Informed Optimization Over Mixtures for \\ Ambiguity-Aware Object SLAM
}

\author{Ziqi Lu$^{1,*}$, Qiangqiang Huang$^{1,*}$, Kevin Doherty$^{1}$, and John Leonard$^{1}$
\thanks{This work was supported by ONR MURI grant N00014-19-1-2571 and ONR grant N00014-18-1-2832.}
\thanks{$^{1}$ Computer Science and Artificial Intelligence Laboratory (CSAIL), Massachusetts Institute of Technology (MIT), Cambridge, MA 02139. \texttt{\{ziqilu, hqq, kdoherty, jleonard\}@mit.edu}. * Equal contributors.}
}

\begin{document}

\maketitle
\thispagestyle{empty}
\pagestyle{empty}

\begin{abstract}
Building object-level maps can facilitate robot-environment interactions (e.g. planning and manipulation), but
objects could often have multiple probable poses when viewed from a single vantage point, due to symmetry, occlusion or perceptual failures.
A robust object-level simultaneous localization and mapping (object SLAM) algorithm needs to be aware of this \textit{pose ambiguity}.
We propose to maintain and subsequently disambiguate the multiple pose interpretations to gradually recover a globally consistent world representation.
The max-mixtures model is applied to implicitly and efficiently track all pose hypotheses, but the resulting formulation is non-convex, and therefore subject to local optima.
To mitigate this problem, temporally consistent hypotheses are extracted, guiding the optimization into the global optimum.
This consensus-informed inference method is applied online via landmark variable re-initialization within an incremental SLAM framework, iSAM2, for robust real-time performance. 
We demonstrate that this approach improves SLAM performance on both simulated and real object SLAM problems with pose ambiguity.
\end{abstract}

\section{Introduction}

Robot-environment interactions rely heavily on real-time, robust and high-level spatial understanding. 
In mobile manipulation, for instance, the robot is required to reliably self-localize and pinpoint the target object for safe navigation and interaction.
Building a consistent object-level world representation would greatly facilitate the process. 
Without external assistance and abundant visual features, the robot is expected to map the environment using an object-based SLAM system.

Recent progress in machine-learning-based object recognition and pose estimation techniques has spurred the development of object-based SLAM \cite{rosen2021advances}.
Using the learned perception models, the robot gains higher-level environment understanding leveraging semantic and object pose information.
However, ambiguities in object poses, induced by symmetry or occlusions, can cause unexpected uncertainties in pose estimates.
The computer vision community has developed various ambiguity-resolving pose inference algorithms \cite{manhardt2019explaining, sundermeyer2020augmented, okorn2020learning, corona2018pose,grossmann2017fast, marton2018improving, deng2019poserbpf, wen2020se} to address the difficulty.
To best leverage these techniques, a SLAM framework must be capable of incorporating the potentially ambiguous local pose measurements and efficiently recover a robust global representation.
 
\begin{figure}[t]
	\centering
	\includegraphics[width=0.75\columnwidth,trim=0 0 0 25]{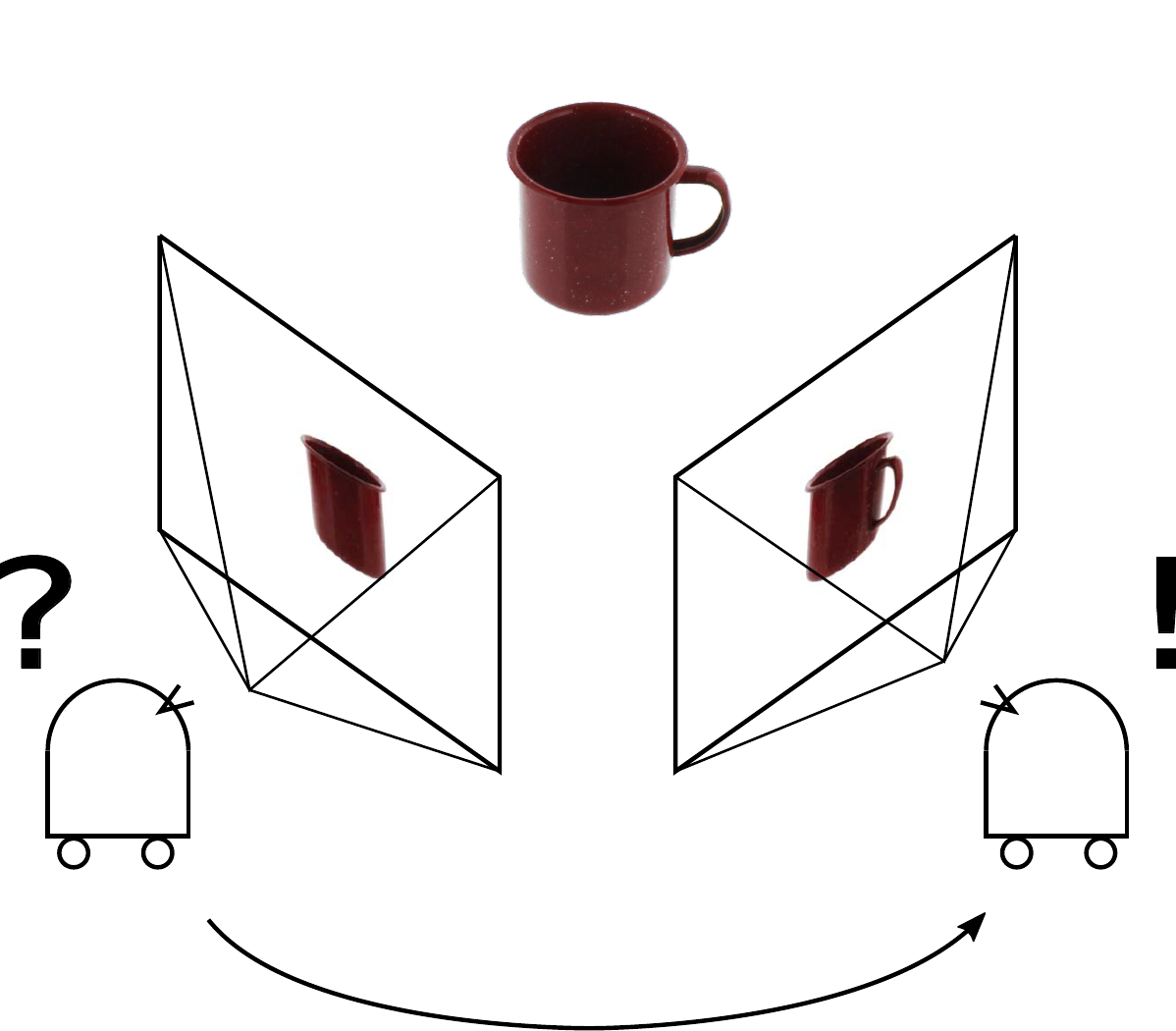}
	\caption{Coffee mug pose ambiguity and disambiguation: A single-shot observation from a handle-occluded viewpoint fails to yield unique robot and mug pose estimates. Disambiguation for robust mug grasping requires a handle-visible viewpoint.}
    \label{fig1}
\end{figure}

For instance, if viewing a coffee mug from a vantage point where the handle is not visible, the mug pose may not be uniquely constrained (Fig.~\ref{fig1}).
An ambiguous observation of a coffee mug would therefore result in multiple, and even infinitely many, possible interpretations of the robot and mug poses.
The ambiguity-oblivious single-hypothesis SLAM method may fail quietly in this circumstance.  

In this work, we use a discrete, multi-hypothesis pose representation to quantify the uncertainties in ambiguous objects' 6D pose measurements.  
We keep track of all the pose hypotheses over time and gradually recover the robot and object poses.
The challenge is that the number of total hypotheses grows exponentially with increasing ambiguous measurements.
An exhaustive global search over the hypothesis space entails exorbitant time and memory.

To alleviate the complexity, we represent all hypotheses as Gaussian max-mixtures, and cast the problem as a continuous, albeit multi-modal, factor graph optimization \cite{olson2013inference}.
High computational efficiency is achieved with Gaussian-preserving max-mixtures amenable to nonlinear least-squares optimization. 
However, the initialization-sensitive gradient-based optimizers can be easily trapped in local optima without global awareness of modes in the posterior. 

We propose a consensus-based heuristic to augment local optimization (iSAM2 \cite{kaess2012isam2}) with global understanding of the dominant pose hypothesis. 
The most consistent set of object pose estimates are extracted from measurements to inform optimization of the dominant mode up to now. 
Once the current mode loses superiority, we perform low-overhead ``dynamic re-initialization (re-init)" to change landmark initialization and restart the optimization inside a new dominant mode.  
This procedure improves robustness of the efficient optimization over mixtures with the flexibility of adjusting mode selections.

The proposed algorithm provides a robust, real-time  solution for robot localization and mapping in feature-sparse, ambiguity-rich environments. 
Two SLAM experiments with ambiguous objects are conducted to demonstrate its improved estimation performance.

\section{Related Work}

\subsection{Ambiguity-Aware Pose Estimation}
Object pose estimation considering symmetry or ambiguity is becoming increasingly popular in computer vision and robotics.
A natural and ordinary idea is to represent symmetry in an unambiguous manner (e.g. \cite{rad2017bb8,kehl2017ssd}).
Rad and Lepetit \cite{rad2017bb8} restricts the pose of a rotationally symmetric object within a unambiguous range.
Any pose measurement beyond the range is returned as its counterpart.

Another popular class of methods make multi-hypothesis pose predictions via view matching.
The visual appearances of an object from arbitrary viewpoints are compared against the captured image.
Multiple poses are returned when there exist one-to-many view-pose correspondences.
Manhardt {\it et al.} \cite{manhardt2019explaining} retrieve 6D poses for ambiguous objects based on learned multi-hypothesis view-pose mapping.
Sundermeyer {\it et al.} \cite{sundermeyer2020augmented} encode visual appearances using an auto-encoder and perform similarity checks in the feature space.

Other ambiguity resolving strategies incorporate regressing pose distributions \cite{okorn2020learning} and 
training symmetry predictors \cite{corona2018pose}. Beyond that, multi-shot pose estimators are developed to gradually recover ambiguous object poses. 
The predicted poses are tracked and updated from successive observations by non-Gaussian filters \cite{grossmann2017fast,marton2018improving, deng2019poserbpf} or a learned tracking network \cite{wen2020se}.

Our primary interest is to tolerate and utilize the local, potentially ambiguous, object pose measurements to build a globally consistent world representation.

\subsection{Multi-Modal SLAM Inference}

We review the three most relevant methods for multi-modal inference under measurement ambiguity.

A natural solution would be to select the most likely hypothesis for each ambiguous measurement and solve an uni-modal inference problem, which we refer to as single-hypothesis method (e.g. nearest neighbor for data association \cite{bar1990tracking}). With the loss of information in discarded hypotheses, it can fail quietly with noise- or outlier-corrupted measurements.

Multi-hypothesis tracking (MHT) based methods\cite{kronhamn1998bearings, huang2013analytically, hsiao2019mh} maintain most probable hypotheses and solve the corresponding set of uni-modal inference problems. B-iMAP \cite{huang2013analytically} is the first smoothing-based incremental MHT SLAM algorithm, where probable hypotheses are selected analytically. MH-iSAM2 \cite{hsiao2019mh} utilizes the Bayes tree and Hypo-tree to perform efficient incremental multi-modal inference with rule-based hypothesis pruning. MHT-based methods provide multiple state estimations including the temporary optimum. However, the rapid hypothesis accumulation necessitates hypothesis pruning. Designing an optimality-guaranteed pruning strategy still remains challenging.

Max-mixtures model \cite{olson2013inference} provides an efficient solution to point estimates in multi-modal inference. However, optimization over max-mixtures can be trapped in a local minimum with a poor initialization. Wang and Olson \cite{wang2014robust} propose to solve the max-mixtures model with stochastic gradient decent (SGD) and obtain increased robustness to poor initializations on pose graph optimizations. However, SGD is sensitive to learning rate and lacks global understanding of dominant modes. Instead of local search, we use the consensus of pose hypotheses to guide optimization towards the global optimum, to obtain a robust \emph{and} efficient solution to pose ambiguity corrupted SLAM.

\subsection{Object SLAM}
SLAM systems utilizing object 6D pose representations (e.g. \cite{civera2011towards, galvez2016real,salas2013slam++, sucar2020nodeslam}) have great promise for robot-object interactions. 
But the initializations of landmark poses, especially for ambiguous objects is a challenging but under-examined problem.
One way of handling ambiguities in single-shot observations is to delay object registrations until robust 6D pose estimations are available (e.g. \cite{civera2011towards, galvez2016real}).
Another approach is to improve the one-shot pose estimates. 
NodeSLAM \cite{sucar2020nodeslam}, for instance, trains a coffee mug rotation estimator for reliable mug orientation initializations.

Instead of relying on the first measurement to provide satisfactory initialization or delaying object registrations, we enable the flexibility to adjust the landmark initializations during the incremental multi-modal inference process.

\section{Object-Based SLAM with Pose Ambiguities}

We define the object-based SLAM problem as the joint inference of 6D robot poses $ X \triangleq \{ x_t \in \mathrm{SE}(3) \}_{t=1}^{T} $ and object landmark poses $ L \triangleq \{ l_j \in \mathrm{SE}(3)\}_{j=1}^{K}$ from a series of measurements $ Z $, consisting of odometry $Z_o = \{z_{t-1, t} \in \mathrm{SE}(3) \}$ and landmark measurements $Z_l=\{z_{tj}\}$, where
$z_{tj} \in \mathrm{SE}(3)$ denotes the relative pose measurement from $x_t$ to $l_j$.
Each object in the world is assumed to be static.
Without measurement ambiguity, the inference can be modeled as a maximum {\it a posteriori} (MAP) estimation problem:
\begin{equation}\label{eq1}
\hat{X}, \hat{L} =  \argmax_{X,L} p(X,L \mid Z)
\end{equation}

In the real world, however, a robot could inevitably fail to uniquely perceive the 6D pose of an object.
Ambiguous objects such as a coffee mug with handle occluded (Fig. \ref{fig1}), or a centrally symmetric playing card don't have unique pose representations.
Furthermore, the perception models can also output highly-uncertain or redundant pose predictions due to algorithmic failures or challenging scenes.

Instead of assuming perfect measurements, we use multiple discrete pose hypotheses to represent all possibilities.
A symmetric playing card, for example,
possesses $ N=2 $ pose hypotheses in a single-shot observation. 
Na\"ively selecting a \emph{single} hypothesis can lead to robot localization errors, since opposing viewpoints yield identical card appearance. Consequently, it is necessary to fully represent the two visually indistinguishable poses.

Therefore, we also have to infer the true hypothesis in each ambiguous measurement. A discrete ``hypothesis decision" variable $ H = \left\lbrace h_{tj}\right\rbrace $ is introduced to the optimization.
$ h_{tj} = i $ $(i=1\sim N)$ indicates that the $ i $-th hypothesis in measurement $ z_{tj} $ is the true pose for landmark $ l_j $. Accordingly, the full MAP estimation can be formulated as:

\begin{equation}\label{eq2}
\hat{X},\hat{L},\hat{H} = \argmax_{X, L, H} p(X, L, H \mid Z)
\end{equation}

\begin{figure}[t]
\centering
\includegraphics[width=1.0\linewidth]{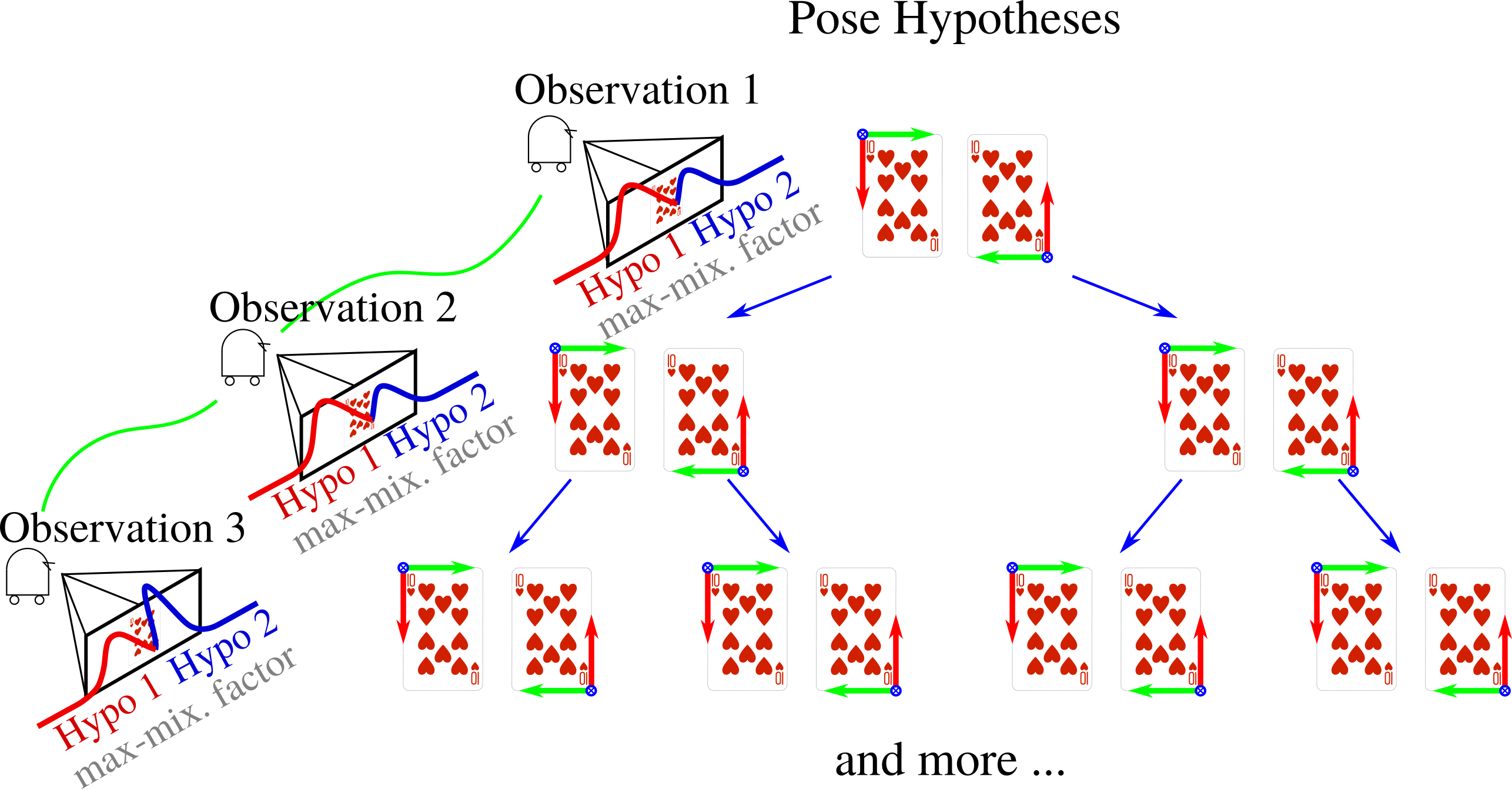}
\caption{Hypothesis explosion. The symmetric playing card has 2 pose hypotheses in each observation. $ M $ observations leads to $ \mathcal{O}(2^M)$ hypotheses in total. We use max-mixture factors to implicitly represent the multiple hypotheses in a measurement as Gaussian mixtures.}
\label{fig2}
\end{figure}

Unfortunately, the combinations of possible hypothesis assignments grow exponentially with the number of ambiguous measurements during navigation (Fig.~\ref{fig2}).
An exhaustive search over the exploding hypothesis space imposes great computational cost. 
We propose to implicitly model the multi-modal uncertainty as Gaussian max-mixtures.

\section{Max-Mixtures Method with Dynamic Re-Init}
\subsection{Max-Mixtures Model}
While max-mixtures was initially introduced as an approximation of sum-mixtures, we show that it can be independently derived by variable elimination of the MAP.
With the robot and object landmark poses, $ X, L $, being of key concern in SLAM, we can marginalize out the hypothesis selections $ H $ from \eqref{eq2} with the max-product algorithm:


\begin{equation}\label{eq4}
\hat{X},\hat{L} = \argmax_{X, L} \left[\max_H p(X, L, H \mid Z)\right]
\end{equation}

Under the Bayes rule, the joint posterior in \eqref{eq4} can be factored as:
\begin{equation}\label{eq5}
\hat{X},\hat{L} = \argmax_{X, L} p(X)p(L)\left[\max_H p(H)p(Z \mid X, L, H)\right]
\end{equation}
with independent joint priors and the $ H $ irrelevant priors taken out of the $\max$ operator.

Assuming the measurements are independent, the joint probabilities can be decomposed as the product of factors (prior factors on $X, L$ omitted):
\begin{align}\label{eq6}
\hat{X},\hat{L} = \argmax_{X, L}
\prod_{Z_o}\phi(x_{t-1}, x_t) [\max _{H}  \prod_{Z_l} \psi(x_{t}, l_j, h_{tj})]
\end{align}
where $\phi(x_{t-1}, x_t)$ represents odometry factors that are irrelevant to pose hypotheses $ H $, and $\psi(x_{t}, l_j, h_{tj})$ corresponds to landmark pose measurements, which following the joint probabilities in \eqref{eq5} takes the form of: 
\begin{equation}\label{eq7}
\psi(x_{t}, l_j, h_{tj}) = p(h_{tj})p(z_{tj} | x_{t}, l_j, h_{tj})
\end{equation}

Since the hypotheses $h_{tj}$ are conditionally independent from one another given $x_t$ and $l_j$, the $\max$ operator can be pulled inside the product operator in front of $ \psi $ in \eqref{eq6}, yielding the final form of the landmark-measurement factor:
\begin{align}\label{eq8}
\phi(x_{t}, l_j)= \max_{H}\psi= \max_{h_{tj}} p(h_{tj})p(z_{tj} | x_{t}, l_j, h_{tj}).
\end{align}

Assuming the measurement model for each pose hypothesis is Gaussian, we obtain a max-mixture type factor, in which $ p(z_{tj} | x_{t}, l_j, h_{tj}) $ represents the Gaussian observation likelihood for $h_{tj} = 1\sim N$ and $ p(h_{tj}) $ denotes the discrete hypothesis weights.
The $\max$-operator selects the best hypothesis given the latent variables $X, L$.

Finally, the formulation for the max-mixtures model can be expressed as: 
\begin{equation}\label{appeq8}
\hat{X},\hat{L} = \argmax_{X, L}
\prod_{Z_o}\phi(x_{t-1}, x_t) \prod_{Z_l} \phi(x_{t}, l_j)
\end{equation}

With Gaussianity locally preserved by the selected mixture components in  $\phi(x_{t}, l_j)$, the max-mixtures model is amenable to efficient solution by nonlinear least squares optimization.
In this study, we use the incremental SLAM framework iSAM2 \cite{kaess2012isam2} to solve (\ref{appeq8}) for better scalability.

Meanwhile, after hypothesis decisions, other subdominant modes are still kept alive in the latent space.
The best components will be re-evaluated at each optimization step \cite{olson2013inference}.
Therefore, all the pose hypotheses are implicitly tracked forward and the exhaustive search over hypothesis space is replaced by the local hypothesis selections.

However, efficiency of the max-mixtures model is achieved at the expense of potential local optimality.
As the number of ambiguous observations increases, modes accumulate rapidly in the posterior distribution. 
Without a good initial value, the solution can easily get trapped in an incorrect mode. 
Therefore, we propose to use consensus over pose hypotheses to directly guide optimization over max-mixtures into the global optimum. 

\subsection{Dynamic Re-initialization}

A good initial value is typically unachievable for an ambiguous landmark variable.
As the first one-shot object observation has multiple hypotheses, it is usually hard to determine which one represents the real pose.
Even worse, a sub-dominant pose hypothesis could become dominant later.
An arbitrary or a temporarily optimal initialization may trap the solution in a local minimum. 
In general, it is difficult for the solution to escape the incorrect mode and converge to the global optimum. 

Therefore, we perform dynamic landmark re-initialization, forcing some variables in the least-square optimization to ``restart" inside the dominant mode. The procedure is summarized in Alg.~\ref{alg1} 

\begin{algorithm}[t]
\caption{Dynamic Re-Init}\label{alg1}
\begin{algorithmic} 
\State Input: new multi-hypothesis relative pose measurements $ \{z_{tj}\}_{i=1}^N $ from robot $x_t$  to ambiguous landmark $ l_j $, corresponding max-mixture factor $ \phi (x_{t}, l_j)$
\If{$l_j$ does not exist in factor graph}
\State 1. Initialize $ l_j $ with $ l^{\text{init}}_{j} $ and add $ \phi $ into factor graph.
\State 2. Create pose cache $ \{l_j\} = \{x_t\oplus z_{tj}\}_{i=1}^N $ for $ l_j $. 
\ElsIf  {$l_j$ exists in factor graph}
\State 1. Identify the most consistent set of poses in $ \{l_j\} $ and compute their average $ \bar{l_j} $ with Alg. \ref{alg2}.
\State 2. Landmark re-init.
\If{dist($ \bar{l_j},  l^{\text{init}}_{j} $) $>$ threshold $ d $}
\State Re-initialize $ l_j $ with $ \bar{l_j} $ using Alg. \ref{alg3}.
\State $ l^{\text{init}}_{j} = \bar{l_j} $
\EndIf
\State 3. Add $ \phi $ into factor graph. 
\State 4. Append $ \{l_j\} $ with $ \{x_t\oplus z_{tj}\}_{i=1}^N $.  
\EndIf
\end{algorithmic}
\end{algorithm}

\subsubsection{Dominant Hypothesis Inference}

\begin{algorithm}[t]
\caption{Outlier Robust Pose Averaging w/ RANSAC}
\begin{algorithmic}
\State Input: pose cache $ \{{l_j}\} $ for landmark $ l_j $
\State Output: average of inlier poses $ \bar{l_j} $
\State 1. Randomly select a subset of poses out of $ \{l_j\} $.
\State 2. Average the selected poses.
\State 3. Find the set of poses $ S $ within distance threshold $r$ to the average pose.
\State 4. Consensus check
\If {$ |S| > $ threshold $ s $ } 
\State Accept $ S $ as inliers and return their average $ \bar{l_j} $. 
\Else 
\State Repeat from Step 1 until maximum number of iterations is reached.
\EndIf
\end{algorithmic}\label{alg2}
\end{algorithm}

We infer the dominant pose hypothesis for each landmark to inform the optimization of the globally consistent mode.

\begin{figure}[t!]
	\centering
	\includegraphics[width=0.99\columnwidth]{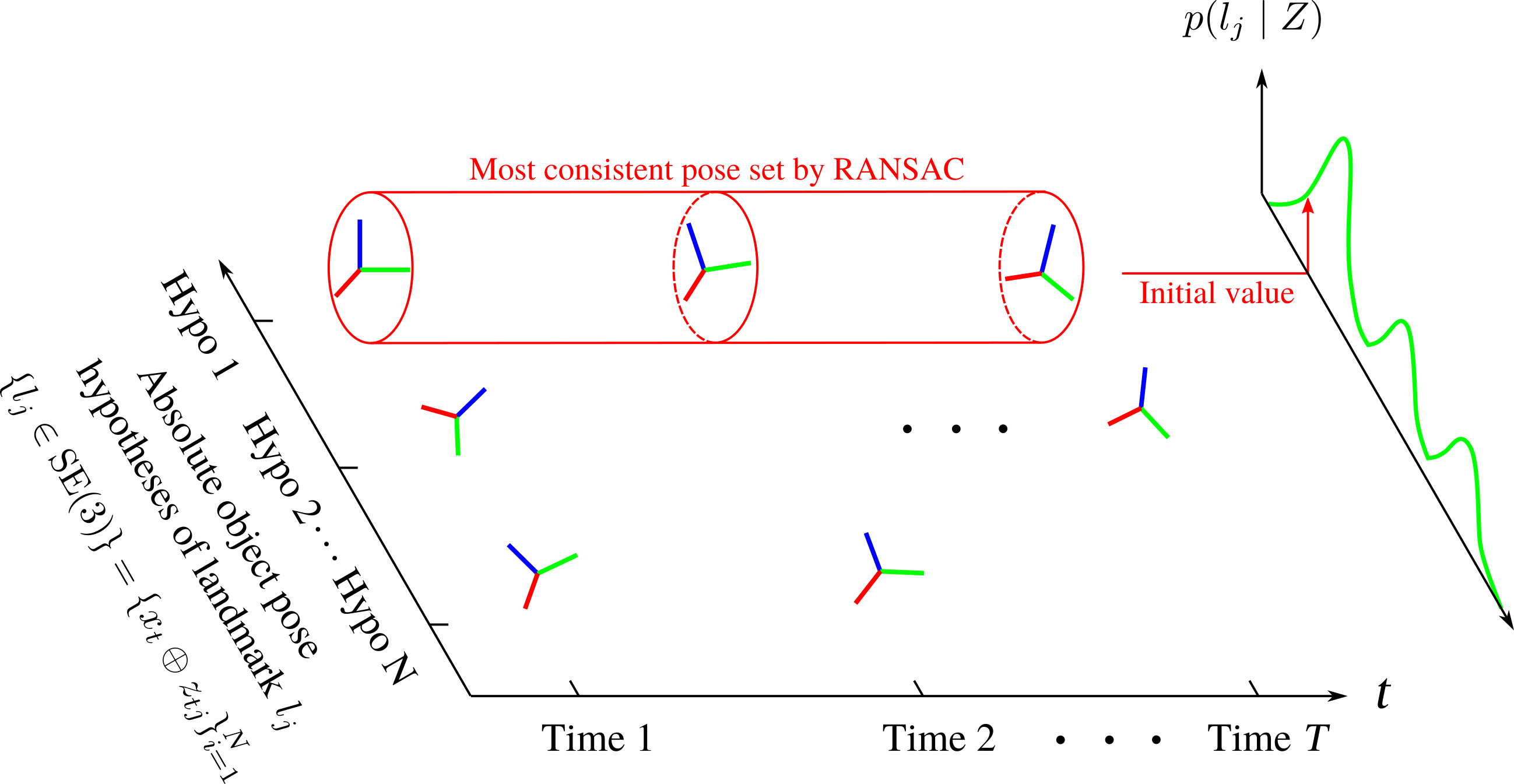}
	\caption{Outlier-robust pose averaging to infer the dominant hypothesis of an ambiguous landmark}
	\label{fig3}
\end{figure}

Since the absolute (world frame) poses for static landmarks are time-invariant, only true pose hypotheses appear consistently in measurements.
False and outlier hypotheses typically exist unsteadily over time. 
Hence, the true hypotheses will gradually build up a leading cluster in our cache of poses $\{l_j\}$.
We extract this consistent set of pose hypotheses and average them to obtain the initial value.
An outlier-robust pose averaging (Fig.~\ref{fig3}, Alg.~\ref{alg2}) method is employed to separate the dominant from sub-dominant hypotheses.
If there's a significant change in the average pose, landmark re-init (Fig.~\ref{fig4}, Alg.~\ref{alg3}) is triggered to guide the optimization into the new dominant mode.

To minimize the parameter tuning efforts, we determine the re-init and RANSAC distance thresholds ($d$, $r$ in Algs.~\ref{alg1}-\ref{alg2}) according to the mutual distances between the hypotheses in multi-hypothesis measurements $ \{z_{tj}\}_{i=1}^N $, which indicates the true-to-false hypothesis distances.
The minimum mutual distance is used to compute and update these thresholds, such that $\text{tol.} < d,r < \min\left[\text{dist}\left(\{z_{tj}\}\right)\right]$.

\subsubsection{Landmark Re-initialization}
As mentioned above, the SLAM variables are estimated using the iSAM2 algorithm \cite{kaess2012isam2}. 
In iSAM2, a landmark variable is initialized at its first observation.
Unfortunately, a ready-made one-stop option to adjust the initialization is not available.
A new function is required to fix incorrect mode selections for already initialized ambiguous landmarks.

\begin{figure}[htb!]
	\centering
	\includegraphics[width=0.84\columnwidth,trim=40 0 40 30]{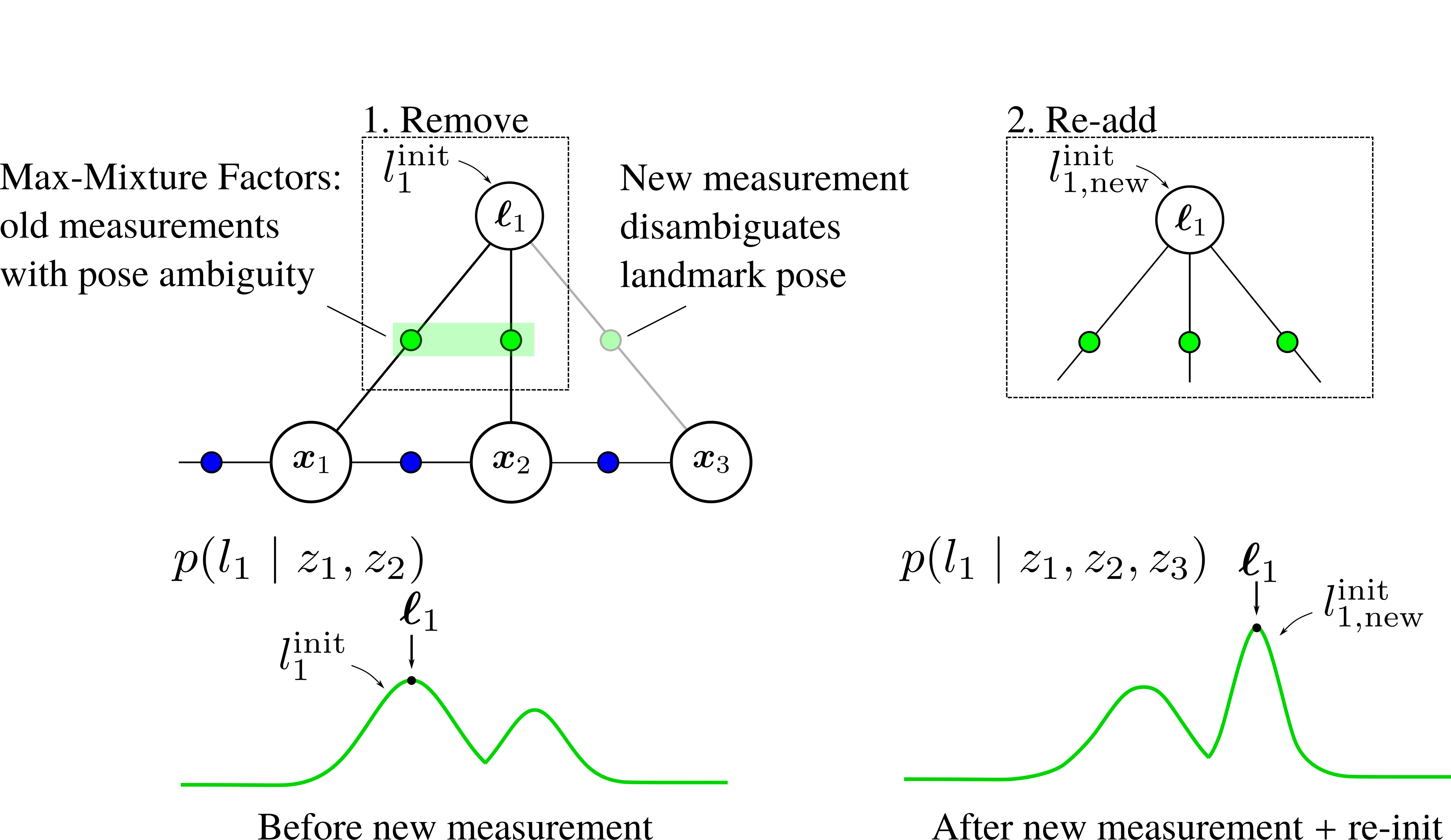}
	\caption{ Landmark re-init by local surgery in a factor graph. 
    At top left is a local part of a factor graph.
    $ l_1 $ is an ambiguous landmark pose variable and $ x_{1,2,3} $ are robot poses.
	Max-mixture factors (green) correspond to the landmark pose measurements $ z_{1,2,3} $, each possessing two pose hypotheses.
	$ l^{\text{init}}_1 $ and $ l_{1,\text{new}}^{\text{init}} $ are previous and new initial values for $ l_1 $.
	The two distributions at bottom are marginal posteriors over $ l_1 $ before and after the new measurement $ z_3 $.
	As $ z_3 $ disambiguates $ l_1 $, we remove and re-add $l_1$ and its neighboring max-mixture factors and re-init it with $ l_{1,\text{new}}^{\text{init}} $.  }
	\label{fig4}
\end{figure}

Fig.~\ref{fig4} visualizes an example for landmark re-init, realized by performing local ``surgery" on the factor graph.
As illustrated, the new measurement $ z_3 $ disambiguates the landmark $ l_1 $, after which the right mode outweighs the left one.
To guide the solution from the left into the right mode, the landmark is re-initialized by removing and re-adding the landmark variable with neighboring factors.
As a result, the optimization is restarted from a new linearization point within the right mode.
Any temporarily optimal mode selections in the old max-mixture factors are corrected.
The solution for $l_1$ converges to the global minimum.

Under the hood, iSAM2 performs incremental local updates for only variables affected by new measurements \cite{kaess2012isam2}.
Provided that the factor graph is not densely connected, the influence of the removing and re-adding steps is rather local. 
Therefore, in most cases, the re-init step does not introduce much overhead, preserving the real-time performance.
This is also validated by a quantitative case study in Appendix A.

\begin{algorithm}[t]
\caption{Landmark Re-Init}
\begin{algorithmic}
\State Input: factors $ \{\phi(\cdot, l_j)\} $ connected to landmark $ l_j $, new initial value $ l^{\text{init}}_{j}$
\State 1. Remove variable $l_j$ and $\{\phi(\cdot, l_j)\}$ from factor graph.
\State 2. Re-add $ l_j $ and $\{\phi(\cdot, l_j)\}$ to factor graph and initialize $ l_j $ with $l^{\text{init}}_{j}$.
\end{algorithmic}\label{alg3}
\end{algorithm}

\section{Experimental Results}
We design two object SLAM experiments\footnote{Please check out the accompanying video for a better result visualization} in ambiguity-rich and feature-sparse environments to test our algorithm .
Playing cards (symmetry-induced ambiguity) and coffee mugs (occlusion-induced ambiguity) are used as test objects.
Our algorithm is developed in C++ using the iSAM2 implementation \cite{kaess2012isam2} in GTSAM library \cite{dellaert2012factor}.
We use ROS \cite{quigley2009ros} for data collection and post-processing. 
We run all the tests on a 2.60GHz Intel i7 CPU.

\subsection{Playing Cards SLAM Experiment}
\begin{figure}[htb!]	
	\includegraphics[width=\linewidth]{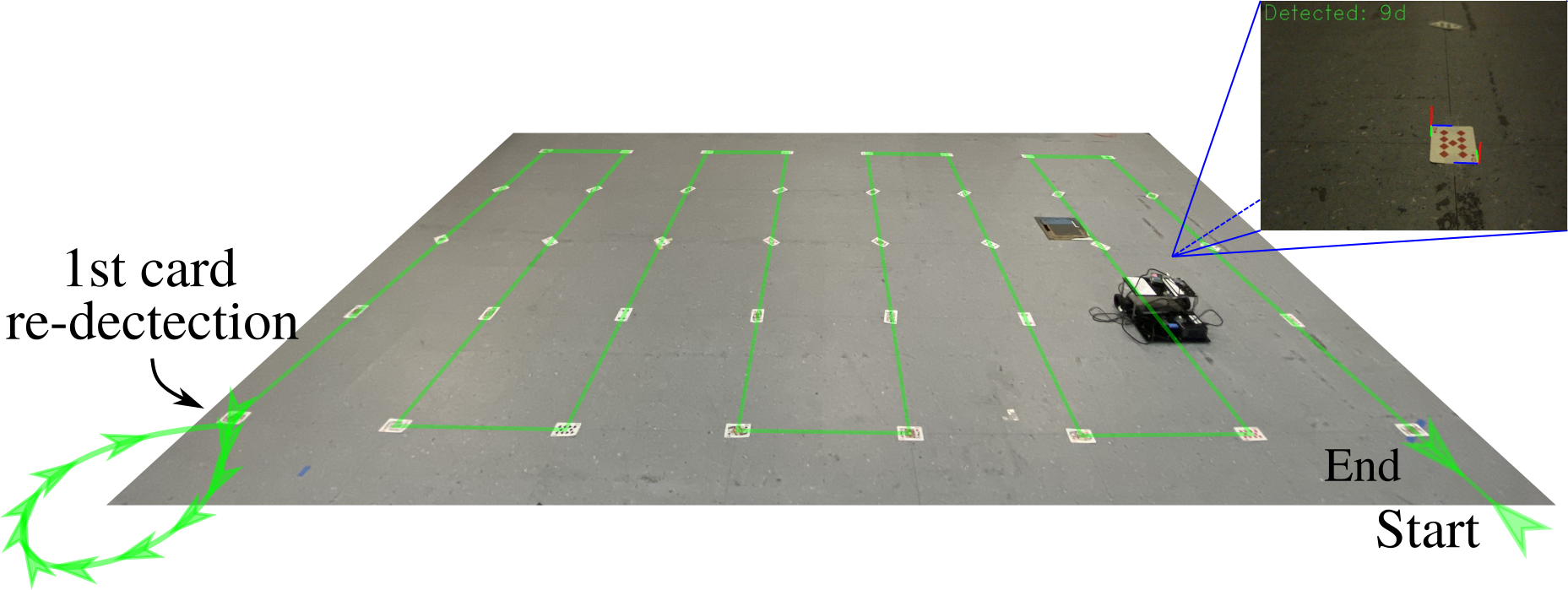}
	\caption{Test field and robot path for the playing cards SLAM experiment. A playing card's two pose hypotheses are estimated from the Blackfly camera view.}
	\label{fig: experimental setup}
\end{figure}

An object SLAM experiment is conducted using playing card landmarks, modelling the symmetric property of ordinary objects. 
As illustrated in Fig.~\ref{fig: experimental setup}, we equip the SwarmRobot~\cite{SwarmRobot} with a forward-pointing ZED camera for visual odometry~\cite{ZED}, and a downward-looking (30$\deg$ to the ground) Blackfly \cite{blackfly} camera for image taking. 
40 playing cards are placed on the ground with a 5$ \times $8 configuration. 
They are drawn from 22 classes (number+suit defines a class), where 4 cards are unique and the rest appear in pairs. 
The robot is remotely controlled to follow a lawnmower-patterned round-trip path.
Each card is first observed from one view angle and revisited later from the opposite. 
Identical card appearances in the two encounters lead to ambiguities in pose estimations.

\begin{figure}[htb!]
    \centering
    \includegraphics[width=0.85\linewidth,trim=90 60 150 90]{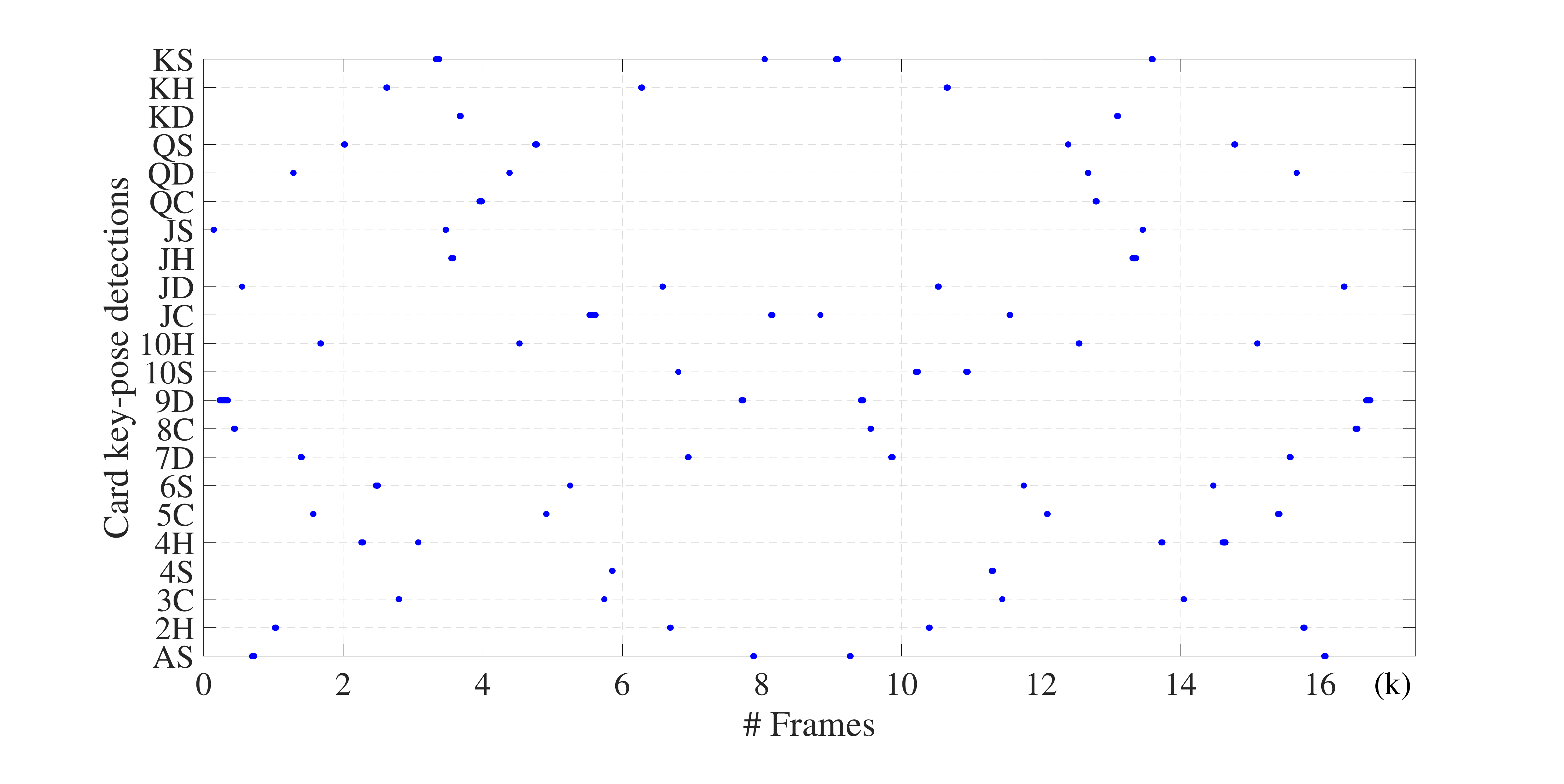}
	\caption{Frames with card key-pose estimations (high-confidence). The sparse structure of this detection matrix reflects the scarcity of environmental features.}
    \label{fig:sparse}
\end{figure}

\begin{figure*}[htb!]
    \centering
    \includegraphics[height=0.3\textwidth,trim=70 0 70 30, clip]{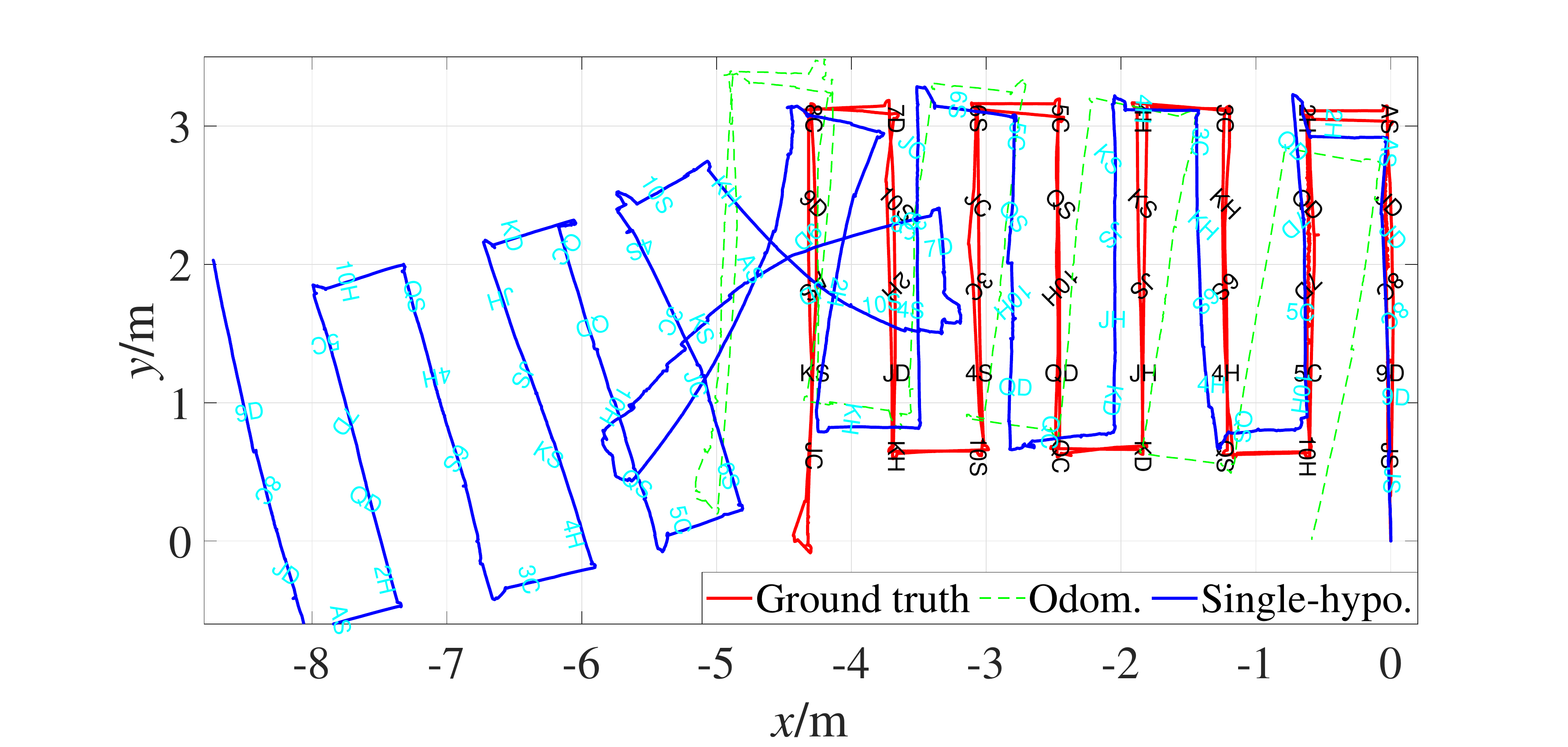}	
    \includegraphics[height=0.3\textwidth,trim=220 0 240 30, clip]{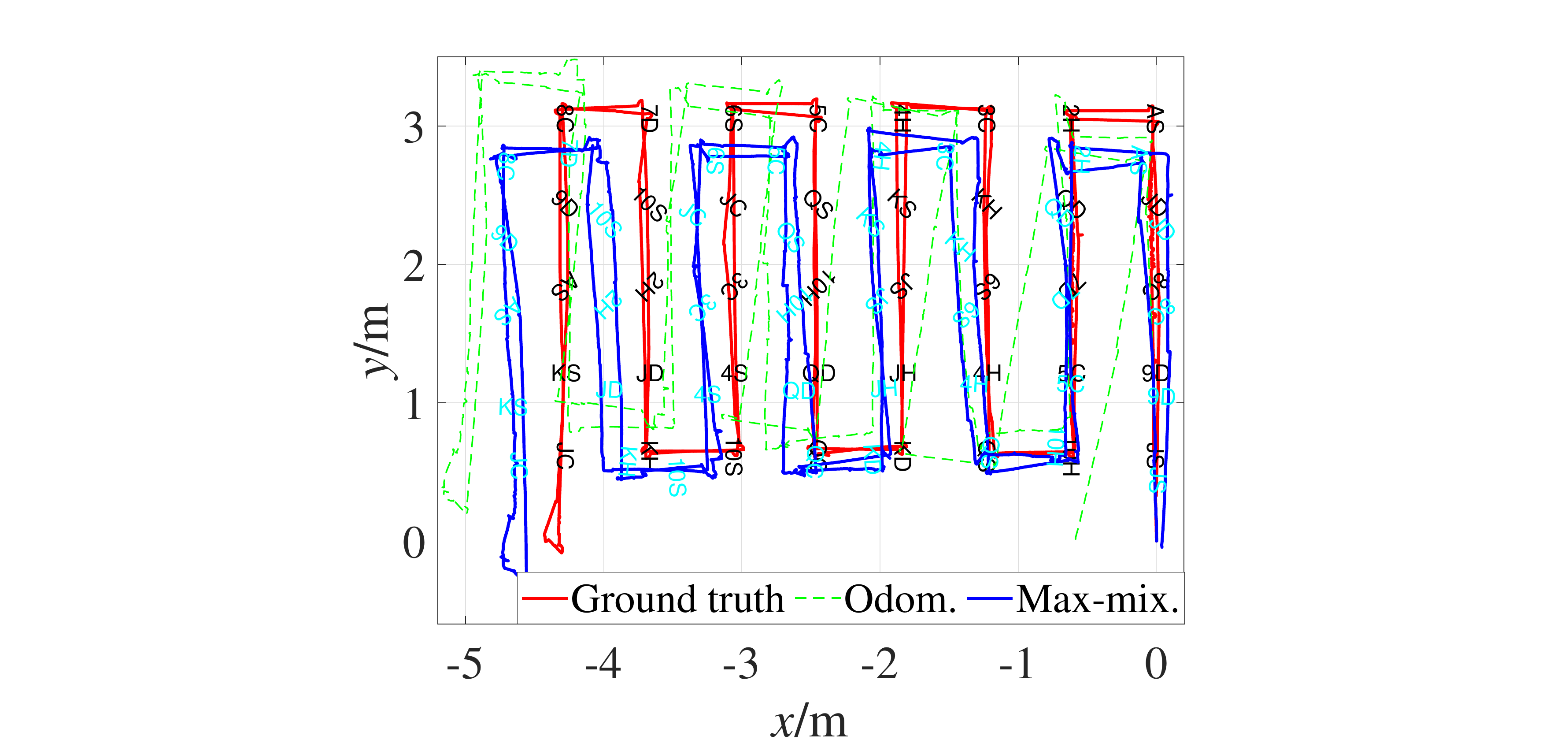}
	\caption{Playing cards SLAM results. Lines are robot trajectories. Black and cyan texts are card pose references and estimations.}
	\label{fig: solutions to the experimental data}
\end{figure*}

\begin{figure}[htb!]
	\centering
	\includegraphics[width=0.9\columnwidth,trim= 45 20 105 85 clip]{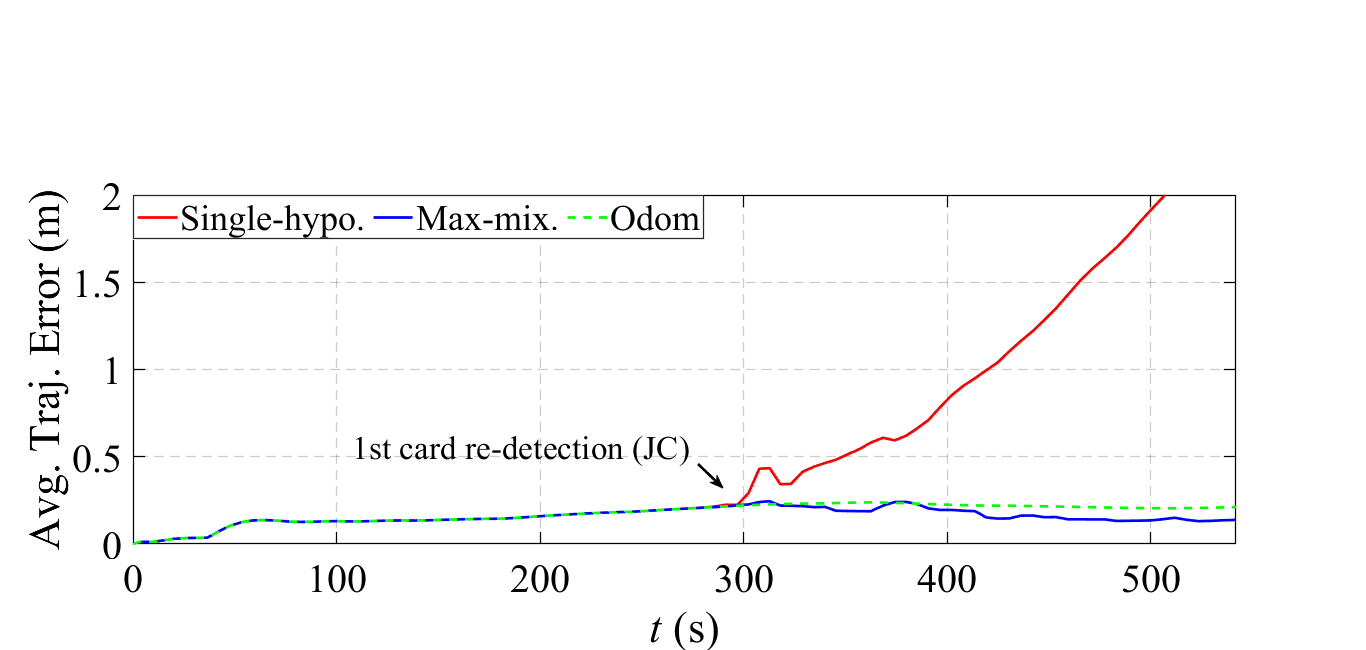}
	\caption{Time evolution of average trajectory errors.}
	\label{fig: estimate error to the experimental data}
\end{figure}

In the roughly 10min long test, 8522 odometric measurements (ZED odometry) and 17369 images are collected. 
The images with cards detected are illustrated in Fig.~\ref{fig: experimental setup}.
We utilize the odometry and estimated relative card poses to gradually recover a world map consisting of robot trajectory and 6D card poses. 
The Vicon MoCap system \cite{vicon} is employed to obtain the ground truth data.

A SIFT feature \cite{lowe2004distinctive} based algorithm is developed using OpenCV for cards identification and pose estimation \cite{poseest}. 
It is able to return the two centrally symmetric card pose hypotheses (Fig.~\ref{fig: experimental setup}).
We design a number of ``key-pose'' criteria to filter out spurious pose measurements and use only key-poses in optimization.
The measurement-card correspondences (data association) are inferred with the classic nearest neighbor approach.
We adopt both the proposed max-mixture algorithm and a single-hypothesis system to solve the optimization. 
The latter incorporates a single-hypothesis estimator, that predicts the most likely card pose, and a uni-modal back-end optimizer. 

As shown in Fig.~\ref{fig: solutions to the experimental data}, the single-hypothesis method fails catastrophically due to oblivion of the pose ambiguity. 
At the first card re-observation, i.e. JC at bottom left of Fig.~\ref{fig: solutions to the experimental data}, the estimator returns near-identical relative orientation compared to the card's first encounter.
However, the true relative orientation is reversed with robot viewing it from an opposite viewpoint.
As a consequence, the inconsistency in JC pose understanding leads to a large deviation in robot localization, which corresponds to the separation point in Fig.~\ref{fig: estimate error to the experimental data}.
Even worse, the localization failure also induces disastrous performance in data associations.
Many cards are registered more than twice in the world map.

On the contrary, the proposed algorithm maintains every pose hypothesis and managed to realize the pose consistency.
It correctly handles the decision making in data associations and pose hypothesis selections.
Loop closures at the JC re-observation and subsequent card revisits constantly adjust the trajectory and minimize estimation errors (Fig.~\ref{fig: estimate error to the experimental data}).
In this feature-scarce and ambiguity-rich environment, as Fig.~\ref{fig:sparse} demonstrates, the max-mixture method still attains good localization and mapping performance.

\subsection{Simulated SLAM Experiment with Coffee Mugs}

\begin{figure}
  \centering
  \includegraphics[width=0.9\linewidth]{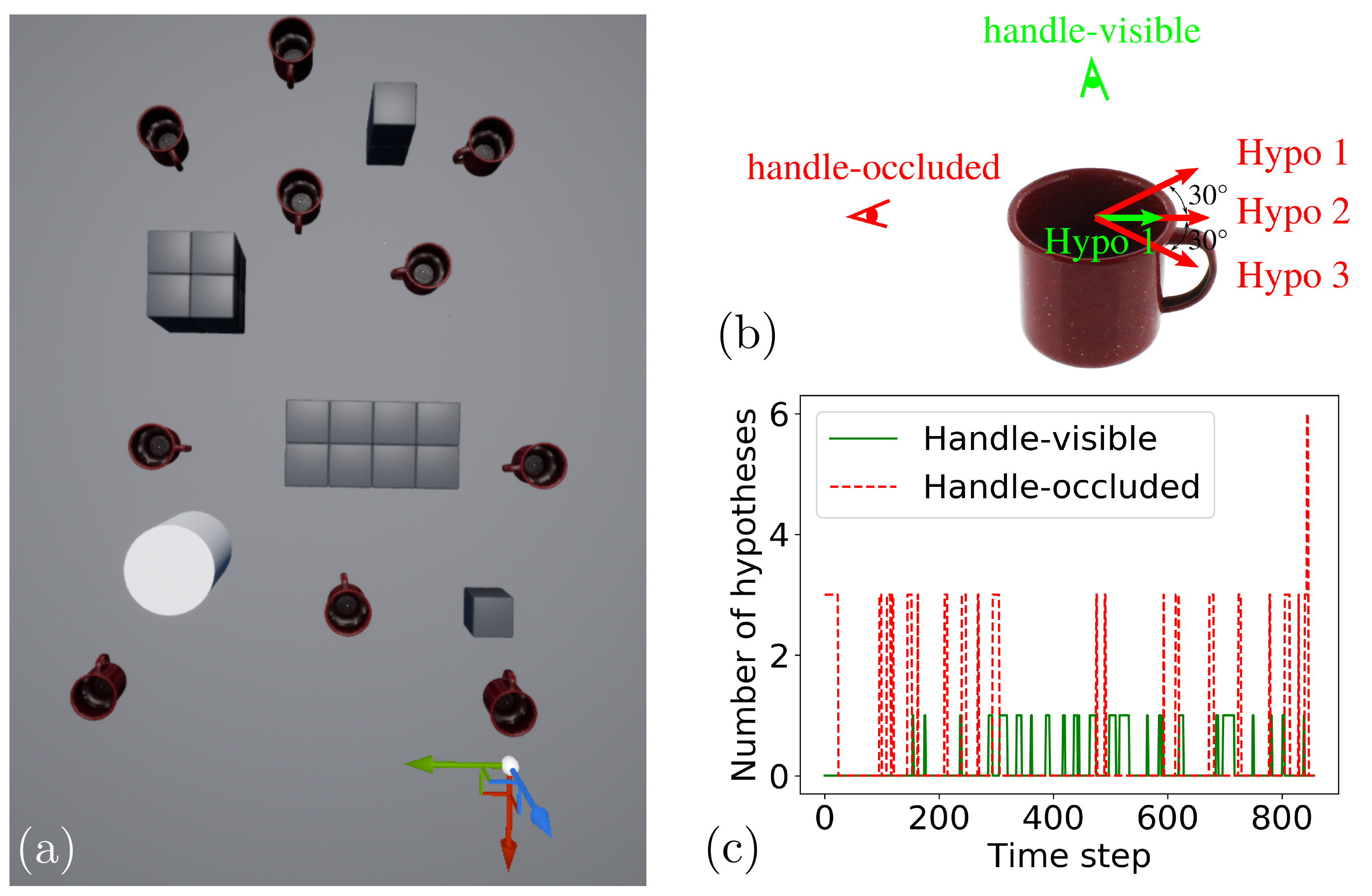}
  \caption{(a) Simulation environment, (b) synthetic poses, and (c) the number of pose hypotheses at each time step. Mugs are scaled for better rendering. The coordinates denote the starting point of the robot. Arrows in (b) represent the directions of mug handle under different hypotheses.}
\label{fig: simulation environment}
\end{figure}

We conduct a simulated SLAM experiment in the Unreal Engine \cite{unrealengine} with 10 coffee mugs as landmarks (Fig.~\ref{fig: simulation environment}(a)).
The mug model is adapted from the YCB data set and has been scaled by 50 folds to match the environment \cite{calli2017yale}. A mobile robot equipped with a monocular camera is integrated in the environment via the AirSim car simulator\cite{shah2018airsim}. The robot odometry, relative poses to the mugs, and camera images are collected via AirSim to estimate the robot trajectory and mug poses. A synthetic pose estimator is created to detect ambiguous poses due to occlusion.
In a handle-occluded case, as shown in Fig.~\ref{fig: simulation environment}(b), a relative pose measurement with three pose hypotheses will be added to the data. Two of the hypotheses are spurious and are derived by corrupting the ground truth pose with pre-determined rotations ($\pm$30 degrees around the vertical axis). The noise for perturbing those pose hypotheses is sampled from a Gaussian distribution in the tangent space about identity pose transformation. The default covariance, $\Sigma$, has diagonal entries of 0.2, 0.2, 0.02, 0.01, 0.01, and 0.01 and zeros on off-diagonal entries (we use the same ordering of Lie algebra components as \cite{barfoot2014associating}). If the handle is visible to the camera, the relative pose measurement will comprise a single hypothesis synthesizing the ground truth pose with noise. There are 857 time steps along the robot trajectory in total. 148 of them are associated with handle-visible detections while 119 of them have handle-occluded detections (see Fig.~\ref{fig: simulation environment}(c)). Note that the small number of detections reflects the sparsity of features in the environment and the large ratio between handle-occluded and handle-visible detections signifies the richness of pose ambiguity.

\begin{figure*}
  \centering
    \includegraphics[width=0.9\linewidth]{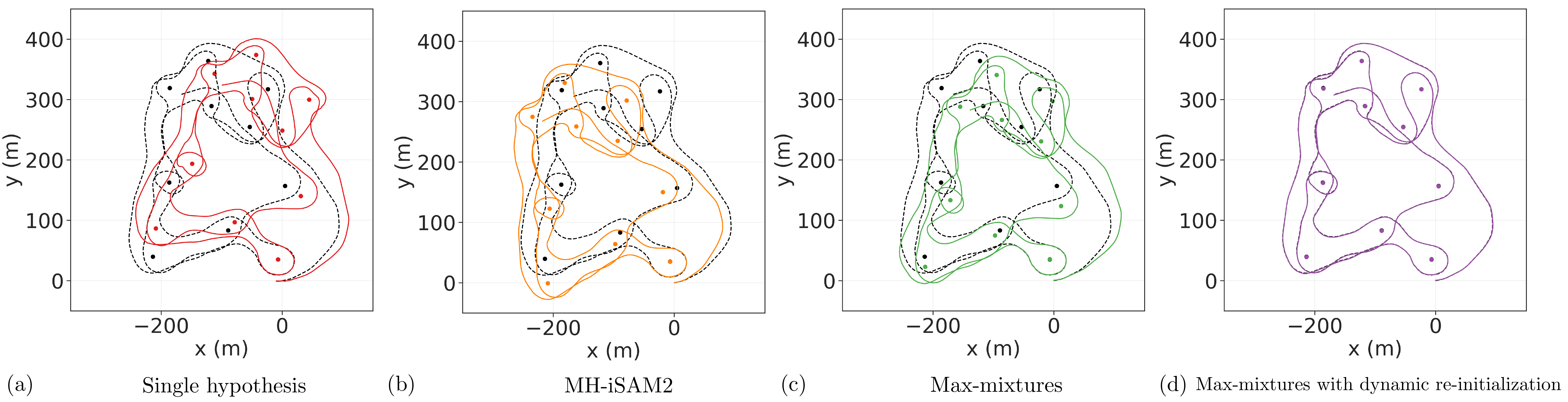}
  \caption{Mug SLAM Estimates. Dots represent mug positions. Reference trajectory and mug positions are in black.}
  \vspace{-15pt}
  \label{fig: estimates by different methods}
\end{figure*}

\begin{figure}[t]
\centering
  \includegraphics[width=0.9\linewidth]{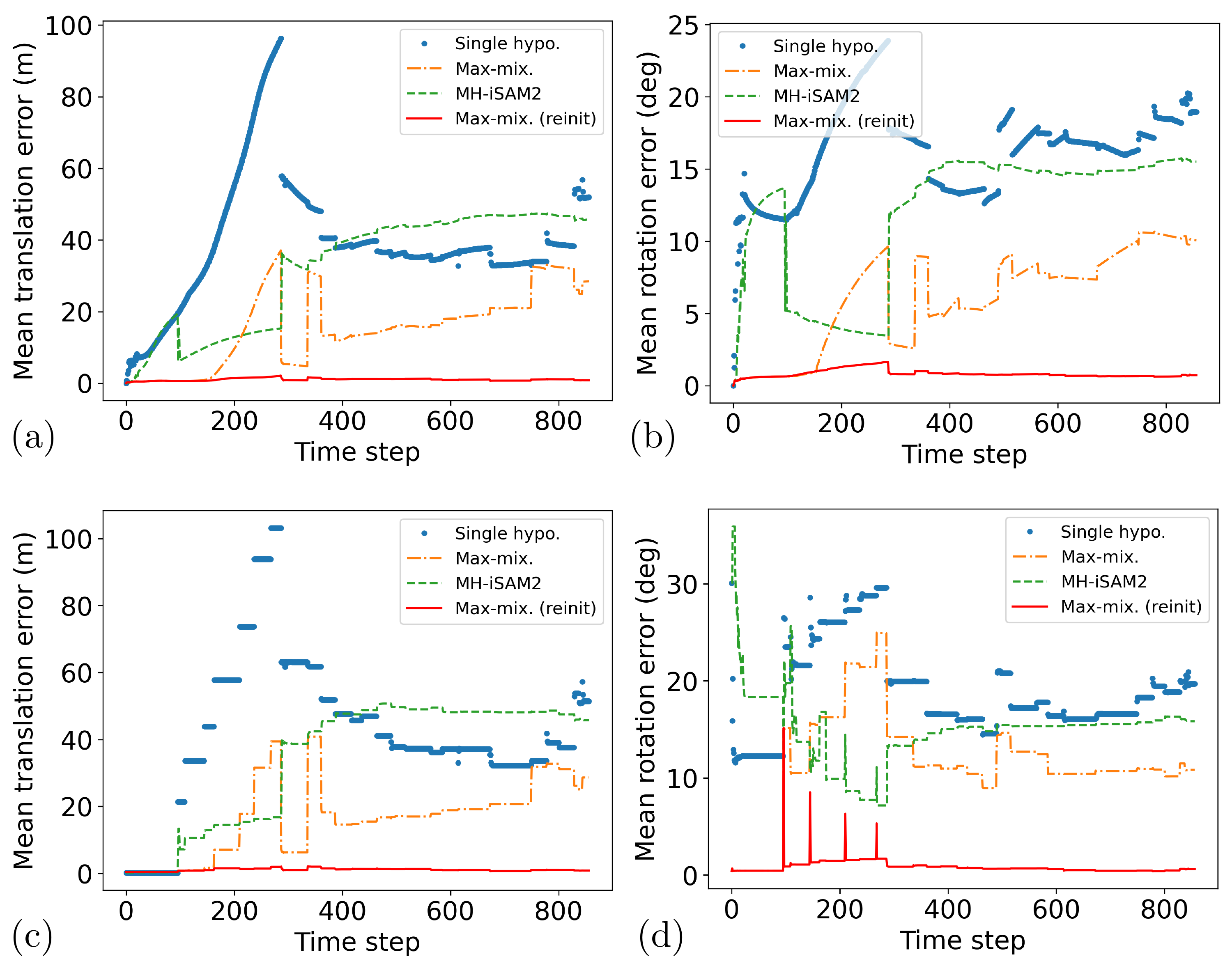}
\caption{Evolution of estimation errors: (a), (b) robot poses; (c), (d) landmark poses.}
\label{fig: estimate errors in the simulation data}
\end{figure}

Fig.~\ref{fig: estimates by different methods} shows the estimates at the final time step by four methods: single-hypothesis, max-mixtures, MH-iSAM2\footnote{We use the MH-iSAM2 implementation at \url{https://bitbucket.org/rpl_cmu/mh-isam2_lib}. The threshold number of tracked hypotheses $n_{\text{limit}}$ is set to 20 (default). The solution from the most probable hypothesis is presented.}, and max-mixtures with dynamic re-initialization. Single-hypothesis refers to permanently admitting a random hypothesis from each measurement with pose ambiguity. It causes large deviations in the trajectory since some measurements are purely represented by spurious hypotheses.
Since the posterior distribution is highly multi-modal in the case considering all hypotheses, the max-mixture estimate can be trapped in local optima due to bad initial values. MH-iSAM2 is supposed to be more robust to initial values as it tracks multiple solutions; however, the hypothesis set for optimal estimate can be pruned even at the very early stage of trajectory, resulting in inaccurate estimates eventually.
Intermediate estimates as the robot proceeds are compared to reveal the evolution of errors. Fig.~\ref{fig: estimate errors in the simulation data} shows that max-mixtures with re-initializing maintains relatively low errors during the whole task. 

\setlength\intextsep{.5\baselineskip plus 3pt minus 2 pt}

To imitate real-world pose detectors with larger errors, we scale the covariance in the default pose perturbation model up and generate more datasets with a larger amount of noise. Table~\ref{table: empirical study} shows estimation errors and runtime.
Max-mixtures with re-initialization performs consistently well in those settings.
Since MH-iSAM2 essentially solves a number of point-estimate problems, it is expected to be slower than others. Although re-initialization in our method is an extra step and causes overhead, it does not necessarily have a big impact on the overall runtime as better initial values can reduce the search time in optimization.

\setlength\textfloatsep{.5\baselineskip plus 3pt minus 2pt}
\begin{table}[h!]
\centering
\begin{tabular}{|c|c|c|c|c|c|c|}
\hline
\multirow{2}{*}{Cov.} & \multirow{2}{*}{Method} & \multicolumn{2}{c|}{MTE (m)} &\multicolumn{2}{c|}{MRE (deg)} & \multirow{2}{*}{Time (sec)}\\
\cline{3-6}
 & &RBT &LMK &RBT &LMK & \\
\hline
\multirow{4}{*}{$5\Sigma$} &SH &42.6 &41.6 &14.8 &12.7 &5.2 \\\cline{2-7}
 &MM &32.3 &32.6 &12.1 &12.6 &5.1 \\\cline{2-7}
 &MH &57.4 &56.1 &14.0 &14.7 &16.0 \\\cline{2-7}
 &\textbf{Ours} &\textbf{10.0} &\textbf{10.3} &\textbf{3.9} &\textbf{3.8} &\textbf{4.8} \\\cline{2-7}
\hline
\multirow{4}{*}{$10\Sigma$} &SH &71.5 &72.1 &23.3 &26.3 &5.7 \\\cline{2-7}
 &MM &40.5 &40.7 &16.5 &15.5 &\textbf{5.3} \\\cline{2-7}
 &MH &73.7 &72.3 &20.1 &20.9 &16.9 \\\cline{2-7}
 &\textbf{Ours} &\textbf{13.9} &\textbf{12.7} &\textbf{7.9} &\textbf{8.0} &7.2 \\\cline{2-7}
\hline
\multirow{4}{*}{$20\Sigma$} &SH &85.4 &86.1 &27.3 &30.0 &7.6 \\\cline{2-7}
 &MM &52.9 &54.8 &18.8 &15.7 &\textbf{6.1} \\\cline{2-7}
 &MH &105.0 &104.8 &27.6 &26.6 &16.1 \\\cline{2-7}
 &\textbf{Ours} &\textbf{26.3} &\textbf{25.0} &\textbf{15.0} &\textbf{11.5} &8.3 \\\cline{2-7}
\hline
\end{tabular}
\caption{Empirical study over noise levels of synthetic measurements. Methods include single hypothesis (SH), max-mixtures (MM), MH-iSAM2 (MH), and max-mixtures with dynamic re-initialization (Ours). MTE is the mean translation error at the final time step. MRE stands for mean rotation error. RBT and LMK are short for robot and landmark, respectively.}
\label{table: empirical study}
\end{table}

\section{CONCLUSIONS}

We present an ambiguity-aware object SLAM inference method that can attain improved localization and mapping performance in feature-sparse, ambiguity-rich environments. 
Our experiment shows that considerable error can arise in the point estimate if pose ambiguities are not considered in the SLAM estimation.
The proposed max-mixture method succeeds under unknown data associations and object pose ambiguities in information-scarce scenes. 
We have also shown that multi-modality-induced local convergence can be mitigated by dynamic re-initialization.
Our experiment demonstrates that the heuristic augmented max-mixture method outperforms state-of-the-art ambiguity-resolving SLAM algorithms.

Our future work involves generalizing pose representations for objects possessing continuum pose hypotheses (e.g. mug).
This requires combining pose distribution inference (e.g. \cite{okorn2020learning}) and more general uncertainty quantifications beyond multi-modality.



\section*{APPENDIX}

\subsection{Runtime Performance Case Study of Landmark Re-Init}

Is landmark re-init an over-treatment while a batch optimization step naturally supports full re-initialization?
To address this question, we case study the runtime performance of landmark re-init.
We build a factor graph with 1000 robot pose variables connected in series.
A number of landmark pose variables are connected with them through randomly generated edges. 
The factor generation and variable initialization are randomized. 
We vary the numbers of landmarks and edges to carry out a series of runtime tests.

We solve the optimizations with iSAM2 \cite{kaess2012isam2} and a batch optimizer (Gauss-Newton).
At the last step of iSAM2 optimization, a random landmark variable is re-initialized.
We time both the normal iSAM2 steps and the last re-init iSAM2 step for evaluation.
The last batch optimization step is also timed for a fair comparison with the re-init iSAM2 step.
We repeat the optimizations 10 times for each parameter combination, and average the time costs.

\begin{figure}[htb!]
	\centering
	\includegraphics[width=0.9\columnwidth,trim= 55 5 95 50 clip]{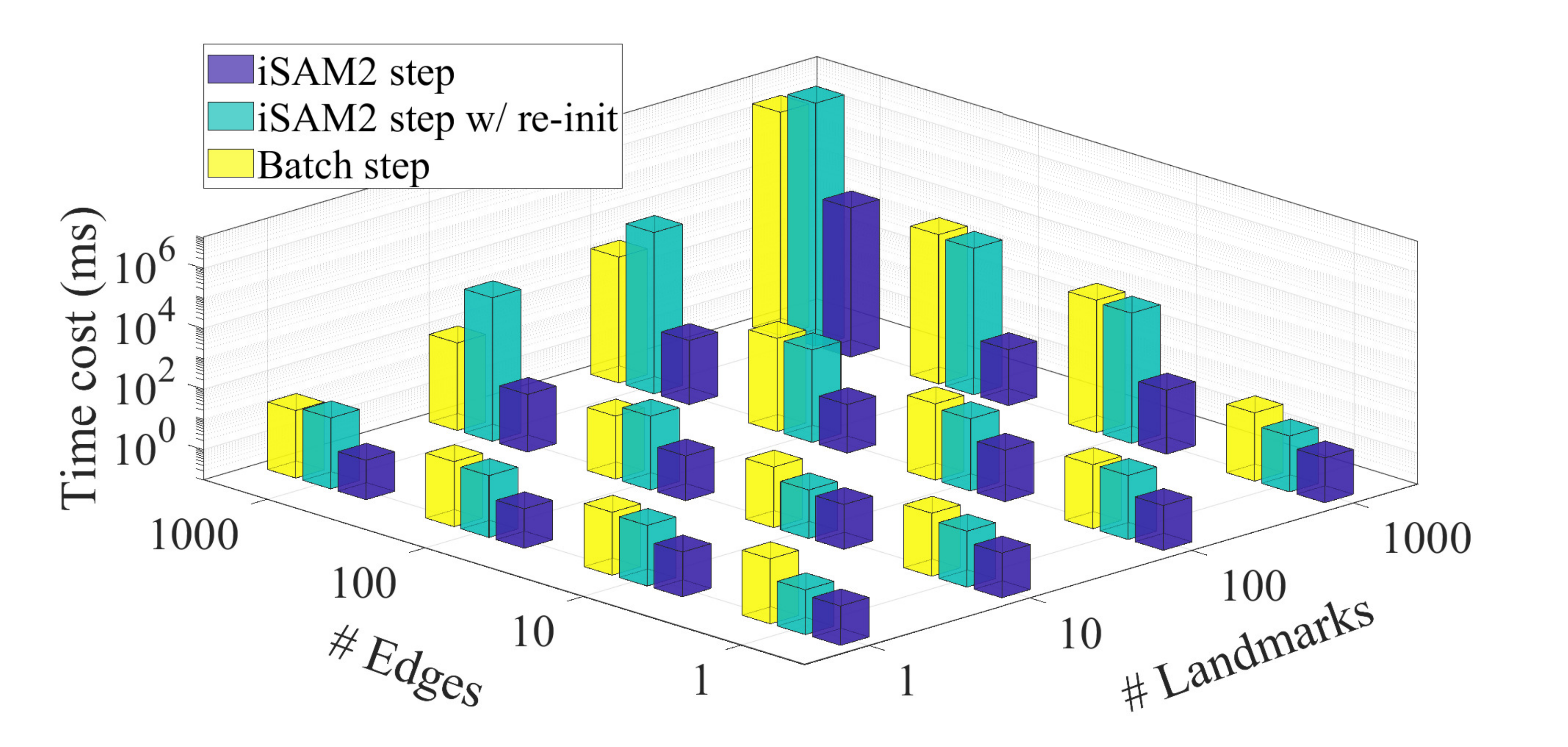} 
	\caption{Runtime performance of landmark re-init evaluated via a factor graph optimization case study.  
    The numbers of landmarks and edges are varied to showcase how the step-wise time cost scales.}
    \label{barplot}
\end{figure}

In summary, the steps in Fig.~\ref{barplot} respectively incorporate:
\begin{itemize}
\item iSAM2 step: factor graph update + estimate computation
\item iSAM2 step w/ re-init: landmark re-init + iSAM2 step
\item batch step: factor graph re-construction + estimate computation
\end{itemize}

\vspace{+.2cm}

We learn from the results that an occasional re-init iSAM2 step does not bring about too much overhead as long as the factor graph is not densely connected.
In the object-based SLAM context we seldom encounter the scenarios where a large number of objects are concurrently observed at all times.
That said, we can conclude that in most cases iSAM2 with re-init is much more efficient than a batch update.

%


\bibliographystyle{IEEEtran}
\bibliography{main}

\addtolength{\textheight}{-12cm}   
	
\end{document}